\documentclass{article}
\usepackage{spconf,amsmath,graphicx}
\usepackage[shortlabels]{enumitem}
\usepackage{hyperref}
\usepackage{multicol}
\usepackage{graphicx} 
\setlist[enumerate]{nosep}

\title{Utilizing Multi-Step Loss for Single Image Reflection Removal}
\name{Abdelrahman Elnenaey, Marwan Torki}
\address{
Department of Computer and Systems Engineering\\
Faculty of Engineering, Alexandria University\\
        Alexandria, Egypt\\
        Knights Lab\\
abdelrhman.elnainay200@gmail.com, mtorki@alexu.edu.eg
        }

\begin{document}
%
\maketitle
\begin{abstract}
Image reflection removal is crucial for restoring image quality. Distorted images can negatively impact tasks like object detection and image segmentation. In this paper, we present a novel approach for image reflection removal using a single image. Instead of focusing on model architecture, we introduce a new training technique that can be generalized to image-to-image problems, with input and output being similar in nature. This technique is embodied in our multi-step loss mechanism, which has proven effective in the reflection removal task. Additionally, we address the scarcity of reflection removal training data by synthesizing a high-quality, non-linear synthetic dataset called RefGAN using Pix2Pix GAN. This dataset significantly enhances the model's ability to learn better patterns for reflection removal. We also utilize a ranged depth map, extracted from the depth estimation of the ambient image, as an auxiliary feature, leveraging its property of lacking depth estimations for reflections. Our approach demonstrates superior performance on the $SIR^2$ benchmark and other real-world datasets, proving its effectiveness by outperforming other state-of-the-art models.


\end{abstract}
\section{Introduction}
\label{sec:intro}
In the realm of computer vision, unwanted reflections in images present significant challenges for applications such as object recognition, image segmentation, scene understanding, and augmented reality. Reflections, which occur due to light interactions with various surfaces, distort visual content and degrade the performance of algorithms that rely on clear images. For example, reflections can obscure features crucial for object detection, introduce ambiguities in scene layout interpretation, and cause misalignment in augmented reality overlays.

Traditional reflection removal methods often require multiple images or sophisticated hardware, making them impractical in scenarios where only a single image is available. 

\begin{figure}[htb]
  \centering
  \centerline{\includegraphics[width=0.99\columnwidth]{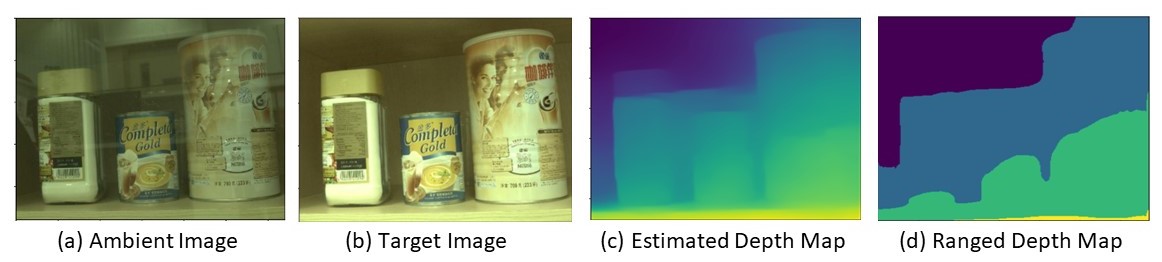}}
    \caption{(c) is the estimated depth map for the ambient image --(a)--, (d) is the ranged depth map with k=4.}
    
\label{depth}
\end{figure}

This paper proposes a novel approach to single-image reflection removal, first introducing a multi-step loss mechanism that offers a more representative training loss and can be generalized to other image-to-image translation tasks with similar input and output characteristics. This mechanism significantly improves model performance and generalization. To address the lack of diverse datasets for training, the paper introduces the RefGAN dataset, synthesized using Pix2Pix GAN, which provides a wide range of reflection patterns and intensities to enhance model robustness. Finally, the approach is further enhanced by using a Ranged Depth Map that excludes reflections from depth estimation, focusing solely on the actual scene content (see Figure \ref{depth}).

The dataset is now available at our GitHub repository.
\href{https://github.com/AbdelrhmanElnenaey/SIRR_MSloss_RefGAN_RDM}{github.com/SIRR\_MSloss\_RefGAN\_RDM}.

The main contributions of this paper can be distilled into three key points:

\begin{itemize}[nosep]
\item We introduce a multi-step loss mechanism, which enhances the learning process by accumulating losses over multiple steps, resulting in improved performance and generalization.
\item We synthesize the RefGAN dataset, which significantly augments the diversity and complexity of the training data, further boosting the robustness and effectiveness of our proposed method.
\item We propose the utilization of a Ranged Depth Map for guiding the reflection removal process, effectively excluding reflections and focusing on the actual scene content.
\end{itemize}

The structure of the paper is as follows: Section 2 reviews related work, Section 3 details the proposed approach and dataset synthesis, Section 4 presents experiments and results, and Section 5 concludes with future research directions.

\begin{figure*}[t]
  \centering
  \centerline{\includegraphics[width=17cm, height=7cm]{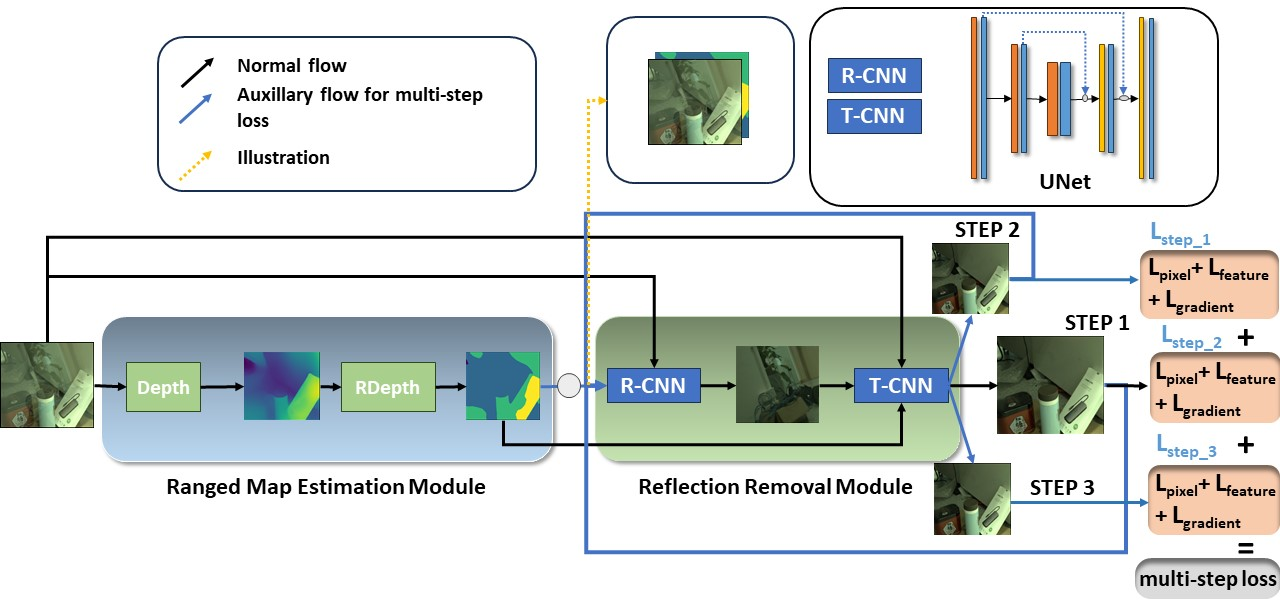}}
    \caption{The architecture of our method. The Depth module estimates the depth of the ambient image. The RDepth module extracts the ranged depth map from the generated depth map. The R-CNN network predicts the reflection. The T-CNN predicts the target. The blue arrows resembles the multi-step loss mechanism during training.}
\label{modelArch}
\end{figure*}

\section{Related Work}
\label{sec:relWork}
\noindent \textbf{Non-machine learning based methods}
Non-machine learning methods include region-aware techniques combining content and gradient information for enhanced precision, as seen in \cite{RegionAware}, and methods that remove reflections and artifacts using non-flash ambient images \cite{argwal}. Another approach leverages depth of field guidance to distinguish in-focus and reflective regions \cite{DepthOfField}.

\noindent \textbf{Single image reflection removal methods}
Single image reflection removal methods began with CeilNet \cite{CeilNet} utilizing deep learning to predict reflection-free images without handcrafted features. Perceptual losses were later introduced to improve the quality of reflection removal \cite{perceptualLosses}. Other advancements include convolutional encoder-decoder networks \cite{EncoderDecoder}, and ERRNet \cite{ErrNet}, which improved performance on misaligned real-world data. Further developments include cascaded refinement methods \cite{IBCLN}, semantic-guided prediction \cite{SGR2N}, and novel architectures with reflection detection and recurrent modules \cite{SOTA}. Recent work introduced self-attention mechanisms and local discrimination loss for more effective reflection separation \cite{SIRR_selfatt}.

In \cite{CloserLook}, a screen-blur combination technique is used for better simulation of reflection characteristics, with SRNet developed to disentangle the reflection and transmission layers. The paper \cite{VIModel} introduces a deep variational inference reflection removal (VIRR) method, which is the first to use this approach, focusing on learning latent distributions for better interpretability.

\noindent\textbf{Flash/no-Flash image reflection removal methods}
Flash/no-Flash methods utilize pairs of flash and ambient images, extracting flash-only images to estimate transmission images without reflections, as demonstrated in \cite{FlashOnly}.

\section{Approach}
\label{sec:appMeth}
Understanding the critical impact of loss functions on the training process, our approach emphasizes the employment of a multi-step loss mechanism to address the shortcomings of conventional loss functions in Single Image Reflection Removal (SIRR). SIRR models often struggle with accurately removing reflections due to the lack of a precise loss function, where the loss function may output a small loss even if the model outputs the input image unchanged. To overcome this, we introduce a multi-step loss function, which iteratively takes the output as input for successive steps, accumulating the loss at each step. Unlike averaging or smoothing the loss across multiple steps, this approach amplifies the loss for incorrectly removed reflections, providing a more representative measure of the model's performance in challenging cases. The model's ability to handle varied reflection intensities is significantly enhanced by observing different reflection ratios in each step for each image, leading to improved overall performance.

\begin{figure}[t]
  \centering
  \centerline{\includegraphics[width=8cm, height=6cm]{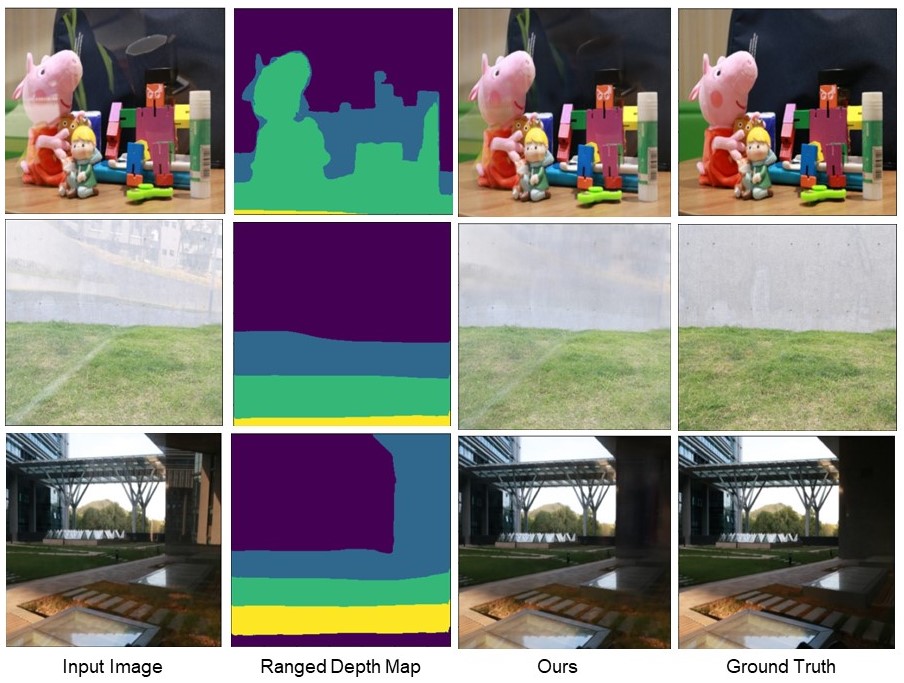}}
    \caption{The model could learn the relations between nearby pixels that fall in the same range in Ranged Depth Map.}
\label{ranged_clarification}
\end{figure}

Our proposed architecture comprises two main modules: the Ranged Map Estimation Module and the Reflection Removal Module. The Ranged Map Estimation Module is responsible for deriving the Ranged Depth Map from the ambient image. We employed a pretrained Midas Small \cite{midas} model to calculate the initial depth map, and subsequently, each pixel in this map is assigned a specific range, forming the Ranged Depth Map. It is important to note that the values in the Ranged Depth Map are non-trainable, ensuring consistent depth estimation throughout the process. Each pixel in the estimated depth map is mapped to a previously constructed range of pixels. The number of ranges is a hyperparameter. Referring to Table ~\ref{n_rangesAbT}, the ideal number of ranges is inferred to be four. The intuition behind using the Ranged Depth Map is that it lacks any estimations for reflections, which helps the model effectively extract the target scene from the ambient image. Moreover, the Ranged Depth Map helps the model to establish relationships between neighboring pixels in the original image that fall within the same range see Figure ~\ref{ranged_clarification}, further enhancing the reflection removal process. The Reflection Removal Module utilizes this Ranged Depth Map as an auxiliary feature for reflection estimation.

Employing an R-CNN architecture, the module incorporates a UNet \cite{UNet} to predict the reflection image. The final stage involves the use of T-CNN, where the estimated reflection, Ranged Depth Map, and the input ambient image are collectively utilized as input. T-CNN, implemented with a UNet \cite{UNet} structure, predicts the target image, providing a comprehensive framework for effective single-image reflection removal. Figure ~\ref{modelArch} illustrates our architecture and the model workflow.

\subsection{Loss Functions}
The training process of the proposed model involves the utilization of three distinct loss functions, each contributing to the refinement of the network.

\textbf{Pixel Loss}, implemented through Mean Squared Error (MSE) loss, is employed to ensure pixel-level accuracy for reflection and target image reconstruction during training.
\begin{align}
L_{\text{pixel}} &= \sum_{T \in D} \left[ \text{$L_{\text{MSE}}$}(T, \hat{T}) + \text{$L_{\text{MSE}}$}(R, \hat{R}) \right]
\label{TotalPixelLossEq}
\end{align}
where T is the target image, and $\hat{T}$ is the predicted target, R is the true reflection, and $\hat{R}$ is the predicted Reflection.

Relying only on the l2-norm loss function frequently results in the loss of high-frequency information.\textbf{Feature Loss} is introduced, applying L1 loss on features extracted from the predicted and actual outputs using VGG19, specifically from conv1\_1 and conv4\_1 layers

\begin{align}
L_{\text{feat}}(\theta) &= \sum_{(I, T) \in D} \sum_{l} \lVert \Phi_l(T) - \Phi_l(f_T(I; \theta)) \rVert_1
\label{FeatLossEq}
\end{align}
where $\phi$ indicates the layer l in the VGG-19 network, T is the true image, and $f_T(I; \theta)$ is the predicted image.

Lastly, \textbf{Gradient Loss}, comparing the gradients of predicted and actual outputs using MSE loss, enhances the model's sensitivity to fine details.
\begin{align}
L_{\text{grad}} &= \sum_{T \in D} \left[ \text{$L_{\text{MSE}}$}(\nabla T, \nabla \hat{T})\right]
\label{TotalPixelLossEq}
\end{align}
The adoption of multi-step loss further refines the learning process by accumulating losses across multiple steps, where the output is iteratively fed as input for successive steps.

The total loss at each step is the sum of the pixel loss, perceptual loss, and gradient loss for predicted output and true output.
\begin{align}
L^t &=  L_{\text{pixel}}^t +  L_{\text{feat}}^t +  L_{\text{grad}}^t
\label{TotalLossPerStep}
\end{align}
\textbf{Multi-Step Loss} is the sum of all losses through all steps.
\begin{equation}
  L_{\text{total}} = \sum_{T\in\mathcal{D}} \sum_{t=1}^{M} \left[L^t(T, \hat{T}_t) \right]
\label{n_stepEq}
\end{equation}

\subsection{RefGAN model}

\begin{figure}[t]
  \centering
  \centerline{\includegraphics[width=8cm, height=6cm]{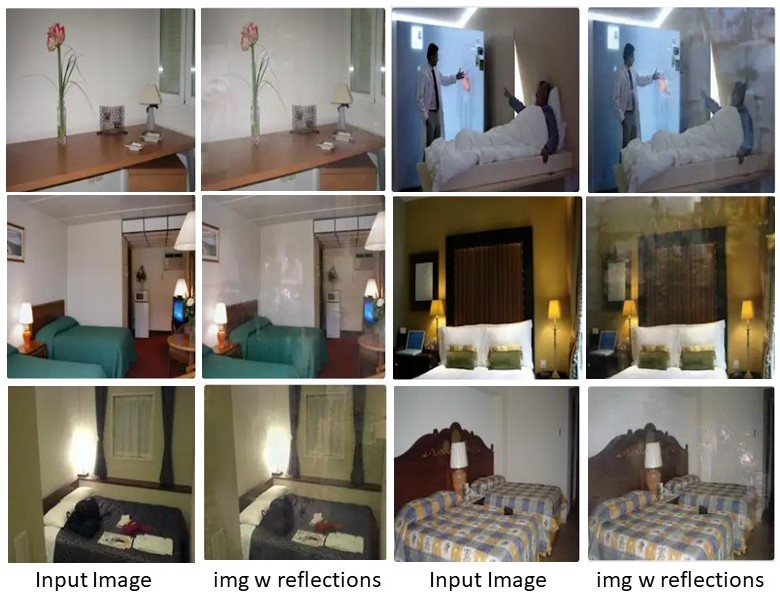}}
    \caption{RefGAN samples generated by pix2pix GAN model.}
\label{refGAN_samples}
\end{figure}

To augment our reflection removal dataset, we constructed the RefGAN model based on the well-known pix2pix GAN.
The pix2pix GAN consists of a UNet generator and a patchwise discriminator. The generator takes a reflection-free image and outputs images with reflections, effectively simulating realistic reflection patterns. The discriminator then evaluates these generated images, distinguishing between realistic images with reflections and non-realistic ones. This process enhances our training dataset by providing a wide range of reflected images, which significantly improves the robustness and performance of our reflection removal model.
Figure ~\ref{refGAN_samples} shows generated samples from RefGAN dataset covering diverse types of reflections (sharpened and foggy reflections).

\begin{figure*}[t]
  \centering
  \centerline{\includegraphics[width=17cm, height=4cm]{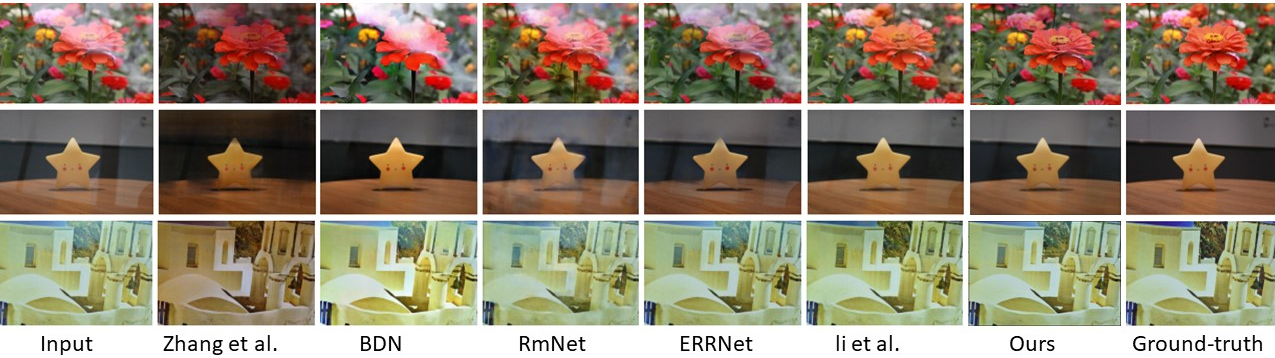}}
    \caption{Contrasting our method against state-of-the-art models on nature dataset (Rows 1-2) and the $SIR^2$ benchmark (Row 3)}
\label{Qual2}
\end{figure*}

\begin{figure}[t]
  \centering
  \centerline{\includegraphics[width=8.3cm, height=2.2cm]{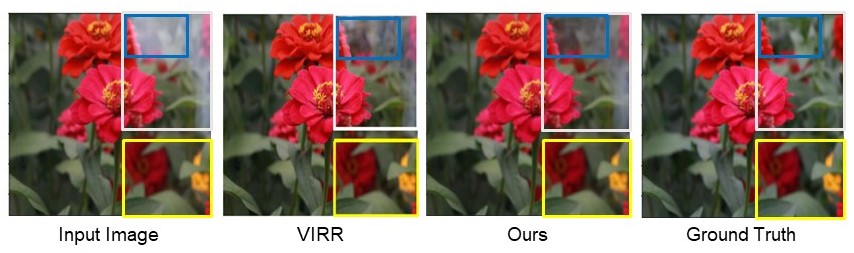}}
    \caption{Contrasting our method against the state-of-the-art VIRR method on nature dataset}
\label{Qual3}
\end{figure}

\section{Experiments}
\label{sec:Exp}
We conducted our experiment on a Nividia Tesla T4 GPU. Our model was trained using a batch size of 1 for 100 epochs. The initial learning rate was $10^{-4}$, and we applied a cosine annealing schedule that starts at $10^{-4}$ and approaches 0 at epoch 100. We used Adam Optimizer and utilized a 2-step loss at first, upgrading to a 3-step loss once the model's performance settled down. Simultaneous training is conducted on the two networks, R-CNN and T-CNN. A single image with a 336 x 336 resolution is inferred in 66.8 ms.

\subsection{Dataset}
\label{sec:Dataset}
The training of our proposed model drew upon a diverse range of datasets to ensure robust performance across various scenarios. Our Training data utilized real-world data from the \cite{FlashOnly} which contains synthesized images from \cite{flashDataset} and \cite{cornDataset} and Nature dataset from \cite{IBCLN}. 

To further enrich the training dataset, we synthesized the RefGAN dataset, comprising 7115 abmient-transmision pair images. The RefGAN dataset was generated using a Pix2Pix GAN \cite{pix2pix} for non-linear synthesis, where the model takes a target image as input and outputs the ambient image containing reflections. Training data for Pix2Pix GAN \cite{pix2pix} exclusively utilized real-world data from \cite{FlashOnly,flashDataset,cornDataset, IBCLN} datasets. 

For validation, we relied on 30 real-world images from \cite{FlashOnly} validation set, ensuring the model's generalization capability. 

Evaluation was conducted on a diverse set of 50 real-world images from \cite{FlashOnly} testing set, 20 real-world images from the Nature dataset \cite{IBCLN}, and the $SIR^2$ benchmark \cite{SIR2Dataset} , which includes three subtesting datasets, namely wild, solid object, and postcard, thereby comprehensively assessing the model's effectiveness in handling a wide array of reflection removal challenges.

\subsection{Quantitative Results}
\label{sec:Res}
A quantitative assessment of our proposed method against state-of-the-art techniques was conducted using state-of-the-art SIRR methods. We report widely accepted metrics such as PSNR, SSIM, and LPIPS. The comparison was conducted on three distinct datasets: the real-world dataset from Flash Only \cite{FlashOnly}, the Nature dataset from IBCLN \cite{IBCLN}, and the $SIR^2$ benchmark \cite{SIR2Dataset}. Table ~\ref{MainExp} consistently demonstrates the superiority of our method. Notably, our approach achieved the best scores, indicated in bold, across all datasets, showcasing its effectiveness in single-image reflection removal. Furthermore, our method secured comparable PSNR scores in the postcard and wild datasets within the $SIR^2$ benchmark. 


\begin{table*}[t]
\centering
\caption{Quantitative comparison shows that our model outperforms all state-of-the-art models on all datasets indicated by bold while achieving comparable PSNR scores on wild and postcard benchmarks.}
\resizebox{\textwidth}{!}{\begin{tabular}{cccccccccccccccc}
\hline
\multicolumn{1}{c}{Data} & \multicolumn{1}{c}{} &  \multicolumn{1}{c}{Kim \cite{Kim}}& \multicolumn{1}{c}{Wei et al \cite{Wei}} & \multicolumn{1}{c}{CeilNet \cite{CeilNet}}& \multicolumn{1}{c}{Zhang et al. \cite{perceptualLosses}} & \multicolumn{1}{c}{BDN \cite{ZAN}} & \multicolumn{1}{c}{RmNet \cite{RmNet}} & \multicolumn{1}{c}{ERRNet \cite{ErrNet}} & \multicolumn{1}{c}{IBCLN \cite{IBCLN}} & \multicolumn{1}{c}{Dong et al. \cite{SOTA}}  & \multicolumn{1}{c}{VIRR \cite{VIModel}}
& \multicolumn{1}{c}{Ours} \\
\hline
real-world(50) & PSNR & 21.67 & 23.89 & -- & 23.76 & 21.41 & -- & -- & \underline{24.31} & -- & -- & \textbf{25.18} \\
 & SSIM  & 0.821 &  0.864 & -- & 0.873 & 0.802 & -- & -- & \underline{0.890} & -- & -- & \textbf{0.904} \\
 & LPIPS & 0.298 & 0.238 & -- & 0.242 & 0.410 & -- & -- & \underline{0.224} & -- & -- & \textbf{0.126}\\
\hline
nature(20)& PSNR & -- & -- & 19.33 & 19.56 & 18.92 & 19.36 & 21.78 & 23.57 & 23.45 & \underline{23.73} & \textbf{24.39} \\
& SSIM  & -- & -- & 0.745 & 0.736 & 0.737 & 0.725 & 0.756 & 0.783 & 0.808 & \underline{0.813} & \textbf{0.906} \\
& LPIPS  & -- & -- &  -- & -- & -- & -- & -- & -- & ---- & -- & \textbf{0.082} \\
\hline
wild (101) & PSNR & -- & -- & 22.14 & 21.52 & 22.34 & 21.98 & 24.16 &  24.71 & 25.73 & \textbf{26.31} & \underline{25.80} \\
& SSIM  & -- & -- & 0.819 & 0.829 & 0.821 & 0.821 & 0.847 & 0.886 & 0.902 & \underline{0.915} & \textbf{0.941}\\
& LPIPS  & -- & -- & -- & -- & -- & -- & -- & -- & ---- & -- & \textbf{0.075}\\ 
\hline
postcard(199)& PSNR & -- & -- & 20.08 & 16.81 & 20.71 & 19.71 & 21.99 & 23.39 & \underline{23.73} & \textbf{24.25} & 23.65 \\
& SSIM  & -- & -- & 0.810 & 0.797 & 0.857 & 0.808 & 0.874 & 0.875 & 0.903 & 
 \underline{0.914} & \textbf{0.931} \\
& LPIPS  & -- & -- & -- & -- & -- & -- & -- & -- & -- & -- &  \textbf{0.099} \\
\hline
object(200)& PSNR & -- & -- & 22.81 & 22.68 & 23.03 & 20.33 & 24.85 & 24.87 & 24.36 & \underline{25.45} & \textbf{25.76}\\
& SSIM  & -- & -- & 0.801 & 0.874 & 0.853 & 0.793 & 0.889 & 0.893 & 0.898 & \underline{0.904} & \textbf{0.943} \\
& LPIPS  & -- & -- & -- & -- & -- & -- & -- & -- & -- & -- & \textbf{0.053} \\

\hline
\end{tabular}}
\label{MainExp}
\end{table*}

\subsection{Qualitative Results}
Figure ~\ref{Qual2} and ~\ref{Qual3} displays the qualitative outcomes for our model compared to cutting-edge ones on the nature dataset and $SIR^2$ benchmark.

\subsection{Ablation Study}
\label{sec:AbStudy}
In our ablation studies, we systematically evaluated key components of our proposed method to better understand their impact on performance. 

Firstly, we compared the use of a pure depth map versus a ranged depth map as an auxiliary feature as shown in Table \ref{depthRangeAb}. While the pure depth map provides pixel-wise distance estimates, it often includes reflections that can mislead the model. In contrast, the ranged depth map excludes these reflections, allowing the model to focus on the actual scene content. Our experiments showed that the ranged depth map consistently produced better results in PSNR and SSIM, indicating its effectiveness in improving reflection removal by better representing the scene and establishing relationships between neighboring pixels as shown in Figure \ref{ranged_clarification}.

\begin{table}[ht]
\centering
\caption{Ranged depth map outperforms depth maps in PSNR and SSIM.}
\begin{tabular}{ccc}
\hline
\multicolumn{1}{c}{} & \multicolumn{1}{c}{depth map} & \multicolumn{1}{c}{ranged depth map} \\
\hline
PSNR & 24.93 & \textbf{25.18} \\
SSIM & 0.908 & \textbf{0.909} \\
LPIPS & \textbf{0.118} & \underline{0.126} \\
\hline
\end{tabular}
\label{depthRangeAb}
\end{table}

Table \ref{refGANAb} shows the PSNR results of training our model with and without the RefGAN dataset on the wild, postcard, and solid objects benchmarks. The results demonstrate that training our model on the RefGAN dataset yields consistently better performance across all benchmarks. This improvement indicates that the model learns more effective patterns for reflection removal when trained with RefGAN, highlighting the dataset's ability to provide high-quality and diverse synthetic data. 

\begin{table}[ht]
\centering
\caption{PSNR results of training our model with and without RefGAN dataset.}
\begin{tabular}{ccc}
\hline
\multicolumn{1}{c}{} & \multicolumn{1}{c}{w/o RefGAN} & \multicolumn{1}{c}{ours} \\
\hline
wild (101) & 25.39 & \textbf{25.80} \\
postcard (199) & 22.57 & \textbf{23.65} \\
object (200) & 24.70 & \textbf{25.76} \\
\hline
\end{tabular}
\label{refGANAb}
\end{table}

Further investigation showed that using 4 ranges in the ranged depth map provided optimal performance. This configuration effectively balances the compromise between convex and concave scenes, allowing the model to handle various image structures, as evidenced by the results in Table ~\ref{n_rangesAbT}.


We investigated the impact of the multi-step loss mechanism, which involves iterative feeding of the output as input for successive steps and accumulating the loss at each step. In our main experiment, we initially leveraged a 2-step loss and then upgraded to a 3-step loss once the model performance had settled down. This upgrade increased the loss after it had stabilized, which in turn boosted the model's learnability and yield best results. Our Experiments detailed in \ref{multi_stepAbT} showed that the 3-step loss, where the output is reused as input two more times, was more effective, enhancing the model's adaptability to varying reflection ratios and providing a better training gradient. In contrast, simpler methods like multiplying a single-step loss by 3 yielded poorer results, highlighting the importance of iterative refinement. This ablation study also involved a simple UNet architecture that takes the ambient image and produces the target image to isolate the effect of the multi-step loss.



\begin{table}[ht]
\centering
\caption{Comparing different Ks performance for 20 epochs on real-world (50). K=4 produces the best results.}
\begin{tabular}{ccccccc}
\hline
\multicolumn{1}{c}{} & \multicolumn{1}{c}{k=2} & \multicolumn{1}{c}{k=3} & \multicolumn{1}{c}{k=4} & \multicolumn{1}{c}{k=5} & \multicolumn{1}{c}{k=6} & \multicolumn{1}{c}{k=7} \\
\hline
PSNR & 24.95  & 24.93 & \textbf{25.06} & 24.92 & 24.78 & 25.01 \\
\hline
\end{tabular}
\label{n_rangesAbT}
\end{table}


\begin{table}[ht]
\centering
\caption{Comparing different multi-step loss configuration on real-world (50).}
\resizebox{\columnwidth}{!}{
\begin{tabular}{cccccc}
\hline
\multicolumn{1}{c}{} & \multicolumn{1}{c}{1-step} & \multicolumn{1}{c}{2-step} & \multicolumn{1}{c}{3x1-step} & \multicolumn{1}{c}{3-step} & \multicolumn{1}{c}{ours}  \\
\hline
PSNR & 24.52 & 24.71 & 23.70 & \underline{24.85} & \textbf{25.18} \\
SSIM & 0.904 & 0.904 & 0.901 & \underline{0.906} & \textbf{0.909} \\
LPIPS & 0.133 & 0.123 & 0.132 & \underline{0.120} & \textbf{0.118} \\
\hline
\end{tabular}
}
\label{multi_stepAbT}
\end{table}

\section{Conclusion}
\label{sec:Conc}

In this paper, we introduce a novel multi-step loss mechanism for single-image reflection removal, which can be generalized to other image-to-image translation tasks where the input and output are similar in nature. This mechanism enhances the learning process, leading to improved model performance and generalization across various scenarios. Additionally, we address the scarcity of reflection removal datasets by creating the RefGAN dataset using Pix2Pix GAN, which significantly boosts the model's robustness and effectiveness in real-world applications. Lastly, we leverage a Ranged Depth Map as an auxiliary feature to focus on the actual scene content by excluding reflections, further enhancing the reflection removal process. Together, these methods have demonstrated superior performance on the $SIR^2$ benchmark and other real-world datasets.

While our study demonstrates the effectiveness of these techniques, it primarily focuses on the loss function, training data quality, and auxiliary input features. Future work could explore integrating these methods with advanced model architectures and more accurate physical models for reflection formation, potentially achieving even greater improvements in performance and generalization across image-to-image tasks.

\fontsize{8.9pt}{9.8pt}\selectfont
\bibliographystyle{IEEEbib}
\bibliography{refs} 

\end{document}